\documentclass[sigconf]{acmart}
\AtBeginDocument{%
  }

\usepackage{amsmath,amsfonts,amsthm}
\usepackage{algorithmic}
\usepackage{algorithm}
\usepackage{array}
\usepackage{textcomp}
\usepackage{booktabs}
\usepackage{stfloats}
\usepackage{multirow}
\usepackage{makecell}
\usepackage{threeparttable}
\usepackage{verbatim}
\usepackage{graphicx}
\usepackage{enumerate}
\usepackage{diagbox}
\usepackage{enumitem}
\usepackage[normalem]{ulem} 
\usepackage{pifont} 
\usepackage[most]{tcolorbox}
\renewcommand\footnotetextcopyrightpermission[1]{}
\newtcolorbox{promptbox}{
colback=gray!5,  
colframe=black!75, 
left=1em, 
right=1em, 
top=1em, 
bottom=1em, 
sharp corners, 
boxrule=1pt 
}

\setlist[itemize]{nosep,leftmargin=*}

\definecolor{myorange}{RGB}{251, 229, 214}
\definecolor{lightred}{RGB}{251,49,153}

\begin{document}
\title{Predicting the Unpredictable: LLM-powered Long-term Chaotic Time Series Forecasting under Short-term Observations}

\author{Yuhang Yao}
\email{2025050423@njupt.edu.cn}
\affiliation{%
  \institution{Nanjing University of Posts and Telecommunications}
  \city{Nanjing}
  \country{China}
}

\author{Bohan Jiang}
\correspondingauthor
\email{bjiang14@asu.edu}
\affiliation{%
  \institution{Arizona State University}
  \city{Tempe}
  \state{Arizona}
  \country{USA}
}

\renewcommand{\shortauthors}{Yao and Jiang}
\begin{abstract}
Chaotic time series forecasting is a challenging task due to its sensitivity to initial conditions and long-term unpredictability. Traditional methods typically rely on sufficient temporal trajectories to learn long-term dynamics, which limits their applicability when only short-term observations are available. While recent Large Language Models (LLMs) have shown great potential for time series forecasting, their temporal representations are not explicitly tailored to the phase-space structure and nonlinear evolution of chaotic systems. To address these issues, we propose PAC-LLM, a \underline{p}hase-space-aware \underline{a}daptive fusion framework for long-term \underline{c}haotic time series forecasting powered by \underline{LLM}s. PAC-LLM leverages learned phase-space features and textual information to fully enable LLM's time series forecasting capacity. In particular, we design an auxiliary feature module and a gated weighting mechanism for multivariate coupling information fusion and selection. Extensive experiments on representative chaotic systems demonstrate that our method outperforms existing fine-tuned and zero-shot baselines in both short-term and long-term predictions. Our ablation study further confirms the effectiveness of each key component in PAC-LLM.
\end{abstract}


\maketitle
\thispagestyle{plain}
\pagestyle{plain}
\section{Introduction}

Chaotic phenomena widely exist in complex systems such as meteorological science \cite{rind1999complexity}, neuroscience \cite{vignesh2025review} and financial modeling \cite{vogl2022controversy}. Predicting chaotic behaviors provides valuable guidance for engineering system scheduling and reliable operation~\cite{liu2022short}. Unlike general time series forecasting tasks, chaotic systems are governed by deterministic rules yet exhibit distinct properties including trajectory divergence, aperiodic evolution, and long-term unpredictability \cite{li2022predicting}. In multivariate scenarios, the future states of the system are further affected by coupling relationships among variables, making prediction errors more likely to accumulate over time~\cite{chattopadhyay2020data}. Therefore, predicting long-term dynamical evolution stands as a core challenge in chaotic time series forecasting. This challenge becomes more pronounced when the available observations are limited. In practical settings, sufficiently long historical observations may be unavailable due to limited sensing duration, newly deployed systems, or online forecasting requirements~\cite{jin2022domain}. This motivates long-term chaotic forecasting from limited recent observations. 

Early research on chaotic time series forecasting mainly adopted traditional machine learning methods, such as Echo State Networks (ESNs)~\cite{jaeger2004harnessing} and Reservoir Computation (RC)~\cite{pathak2018model}. However, their performance often depends on manually designed reservoir structures and network topologies~\cite{viehweg2025systematic}. To better capture the strong nonlinearity and nonstationarity of chaotic sequences, several methods have been introduced to model temporal dependencies with RNNs, TCNs, and Transformers~\cite{shahi2022prediction,cheng2021high,fu2024mixformer}, as well as multivariate coupling structures with 2D-CNNs and GNN variants~\cite{tang2025learning,xiong2024dynamic}.
While recent large language model (LLM)-based methods have shown great proficiency in modeling temporal patterns for general time series forecasting tasks~\cite{xue2023promptcast,liu2024lstprompt,tang2025llm}, their effectiveness on predicting chaotic systems remains underexplored. For example,~\cite{liu2024llms} focused on the zero-shot forecasting performance rather than designing an architecture tailored to chaotic dynamics.

In this paper, we argue that long-term chaotic time-series forecasting is a challenging task that requires recovering local dynamical states from limited (i.e., short-term) observations.
To address this problem, we develop \textbf{PAC-LLM}, a phase-space-aware adaptive fusion framework that jointly leverages the general time series knowledge from LLMs and the learned phase-space features from limited observational data. Instead of directly applying pretrained LLMs to chaotic sequences, PAC-LLM adapts LLM-based forecasting to chaotic dynamics through three key designs. First, PAC-LLM adopts a learnable \textbf{delay-coordinate embedding} to supplement local dynamical state information and construct phase-space-aware representations from short historical observations. Second, a pretrained LLM is used to extract temporal evolution patterns, while multivariate coupling information is introduced as \textbf{complementary auxiliary signals} to refine LLM's representations. Third, rather than naive auxiliary feature concatenations, PAC-LLM introduces a \textbf{gated adaptive weighting mechanism} to control the strength of auxiliary corrections, thereby reducing the risk of noise interference caused by redundant coupling information. Therefore, PAC-LLM considers the contextual representation from pretrained language models with the phase-space structure and variable coupling characteristics of chaotic systems. 

We conduct extensive experiments on representative chaotic systems. The evaluation covers both synthetic and real-world settings. Our results show that PAC-LLM outperforms nine state-of-the-art fine-tuned and zero-shot baselines on both short-term and long-term forecasting. Our ablation studies further illustrate the effectiveness of the core components in PAC-LLM.
In summary, our key contributions are as follows:
\begin{itemize}
	\item \textbf{Task Formulation}: We formulate long-term chaotic forecasting from limited recent observations. This setting addresses real-world scenarios lacking long-term historical records. The forecast horizon is substantially longer than the observed context. We also compare different input lengths to verify that the results do not depend on a single context setting.
	\item \textbf{Framework Development}: We develop an LLM-powered forecasting framework that enhances pretrained sequence modeling with Takens-inspired phase-space representation, statistics-guided modulation, and an auxiliary interaction branch for nonlinear coupling modeling.
	\item \textbf{Comprehensive Evaluation}: We conduct extensive experiments on both synthetic and real-world chaotic systems against representative baselines. The evaluation covers forecasting accuracy, valid prediction time, and long-term dynamical structure preservation. We also perform ablation, robustness, and LLM backbone studies. Experimental results demonstrate the effectiveness and generality of PAC-LLM for chaotic forecasting.
\end{itemize}

\section{Related Work}
\subsection{Chaotic Time Series Forecasting}
Chaotic time series forecasting has evolved from state-space reconstruction and reservoir computing to deep representation learning. Reservoir computing methods map observations into a high-dimensional dynamical space and learn lightweight readout layers~\cite{li2024higher,pathak2018model}, with extensions to spatiotemporal and multi-scale chaotic systems~\cite{srinivasan2022parallel,chattopadhyay2020data}. NVAR~\cite{gauthier2021next} and its variants~\cite{yang2025improved} instead construct explicit nonlinear features from delayed observations, reducing dependence on random reservoirs, although their performance remains sensitive to state construction and hyperparameter settings. Subsequent advances in deep learning have shifted chaotic forecasting toward data-driven representation learning. Models such as LSTM~\cite{vlachas2018data}, GRU~\cite{shahi2022prediction}, and TCN~\cite{dudukcu2023temporal} learn temporal dynamics directly from observed trajectories. Transformer-based methods further capture long-range dependencies~\cite{fu2024mixformer}, while GNN- and CNN-based architectures model inter-variable coupling and spatial correlations in multivariate chaotic systems~\cite{xiong2024dynamic,tang2025learning}.

More recently, LLMs have shown promise for chaotic and dynamical system forecasting. Chronos demonstrates that general time-series foundation models can extract nonlinear predictive patterns from context and preserve attractor statistics in chaotic systems~\cite{zhang2025zero}. DynaMix focuses on context-driven dynamical system reconstruction with long-term statistical preservation, while Panda learns chaotic dynamics representations from large-scale synthetic systems~\cite{hemmer2026true,lai2025panda}. 
However, these efforts relied on sufficient input contexts. Moreover, they varied in terms of training, testing, and evaluation settings, making a fair model comparison difficult. Therefore, we propose a unified setting for studying LLM-powered long-term chaotic forecasting under short-term observations.
\subsection{Large Language Models for Time Series Forecasting}
Recently, LLMs have gained increasing prominence in time series forecasting due to their powerful capabilities in contextual modeling and cross-task generalization. PromptCast transforms time series into textual prompts to bridge time series
and text modalities~\cite{xue2023promptcast}, while Time-LLM maps time series segments into
the representation space of LLMs with auxiliary textual prompts~\cite{jin2024time}. CrossTimeNet further employs a pretrained language model for cross-domain
time series representation learning through discrete tokenization and
self-supervised pretraining~\cite{cheng2025cross}. Subsequent studies introduce
more structured adaptations, including multi-scale decomposition in
LLM-Mixer, cross-modal alignment in TimeCMA, and joint
temporal-semantic modeling in LLM-PS~\cite{kowsher2025llm,liu2025timecma,tang2025llm}. More recently, TimeReasoner
explores slow-thinking LLMs for training-free forecasting through
inference-time multi-step reasoning~\cite{cheng2026can}. However, most existing methods are designed for general time series tasks. They lack dedicated modeling mechanisms to address the unique challenges of chaotic systems, including high sensitivity of local dynamics, complex inter-variable couplings, and rapid accumulation of errors during long-term rolling prediction.\par 
Our proposed method aims to retain the strength of LLMs in long-range contextual modeling, while making the representation learning process better tailored to the dynamical structures of chaotic time series data. This further improves the accuracy and stability of long-term prediction under limited historical observations.

\begin{figure*}[t]
    \centering
    \includegraphics[width=\textwidth]{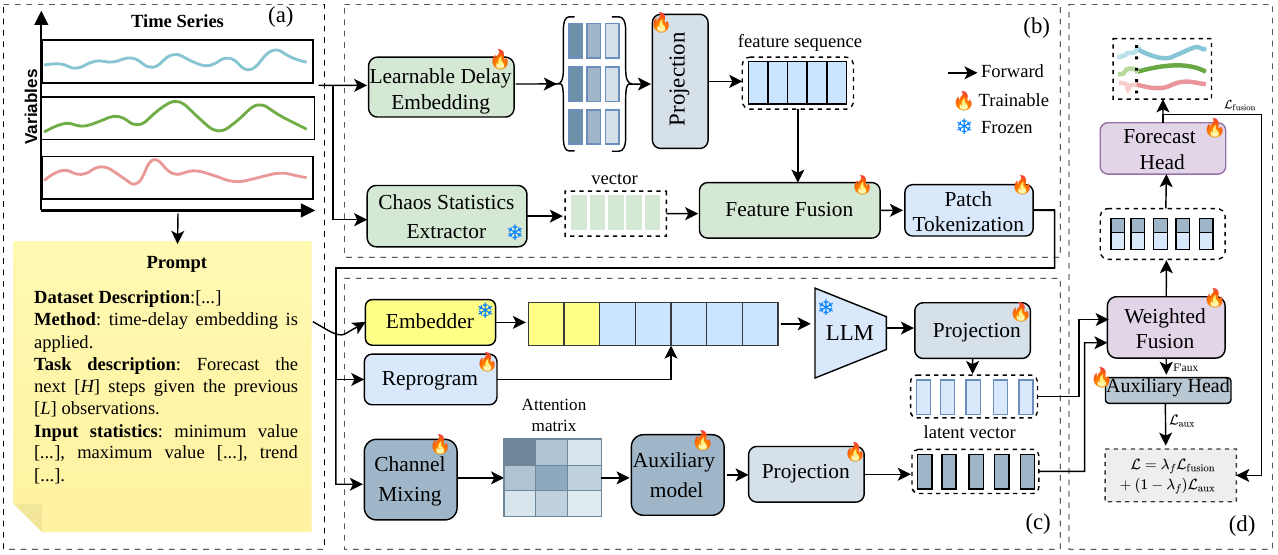}
    \caption{Overview of the PAC-LLM framework. (a) Input multivariate time series and prompt construction. (b) Phase-Space-Aware Representation Learning. (c) Dual-Branch Dynamics Modeling. (d) Weighted Fusion Forecasting with joint loss function. }
    \label{fig:framework}
\end{figure*}
\section{METHODOLOGY}
\noindent\textbf{Task Formulation.} We define the task of forecasting the long-term evolution of chaotic systems based on limited observations as follows. 
Let $\mathbf{X} \in \mathbb{R}^{T \times V}$ denote the historical multivariate observations over a look-back window of length $T$, defined as
\begin{equation}
    \mathbf{X} = [\mathbf{x}_1, \mathbf{x}_2, \dots, \mathbf{x}_T]^\top,
\end{equation}
where each $\mathbf{x}_t \in \mathbb{R}^V$ is the $F$-dimensional observation vector at time step $t$. Here, $V$ denotes the number of observed variables, and $T$ is the available context length.

Given $\mathbf{X}$, the goal is to forecast the future system states over the next $H$ time steps,
\begin{equation}
    \hat{\mathbf{Y}} = [\hat{\mathbf{x}}_{T+1}, \hat{\mathbf{x}}_{T+2}, \dots, \hat{\mathbf{x}}_{T+H}]^\top \in \mathbb{R}^{H \times V},
\end{equation}
where $H$ denotes the forecasting horizon. In our setting, the input context length $T$ is chosen to approximately correspond to \textbf{one Lyapunov time of the synthetic system}, which serves as a characteristic predictability scale in chaotic dynamics. According to the research on long-term prediction of chaotic systems~\cite{zhang2025zero}, we set the maximum forecasting horizon $H$, which covers multiple future Lyapunov times.
\subsection{Overview of the Proposed Framework}
The proposed framework, PAC-LLM, illustrated in Figure~\ref{fig:framework}, consists of four main components:
\begin{itemize}
    \item \textbf{Phase-Space-Aware Representation Learning} (b) transforms raw observations into dynamical representations that are more informative for long-term chaotic forecasting.
    \item \textbf{LLM-powered Temporal Modeling} (the upper part of c) feeds the reprogrammed tokens into the LLM backbone to capture long-range temporal dependencies for future forecasting.
    \item \textbf{Auxiliary Channel Interaction Modeling} (the lower part of c) captures cross-variable dependencies through channel mixing and attention-based auxiliary modeling.
    \item \textbf{Weighted Fusion Forecasting} (d) selectively fuses auxiliary and temporal features for final prediction.
\end{itemize}
\subsection{Phase-Space-Aware Representation Learning}
\noindent \textbf{Learnable Delay Embedding.}
Standard deep learning models typically process time series as sequential data in the time domain, neglecting the underlying topological structure of the dynamical system. According to \textit{Takens' Embedding Theorem}~\cite{takens1981lecture}, the state space of a dynamical system can be topologically reconstructed from a scalar time series and its time-delayed copies.

Inspired by the delay-coordinate idea behind Takens' embedding theorem, we introduce a learnable delay embedding module to construct phase-space-aware representations from short observations. Given a historical observation
\begin{equation}
	X = [x_1, x_2, \dots, x_T] \in \mathbb{R}^{T \times V},
\end{equation}
where $x_t \in \mathbb{R}^{V}$ denotes the multivariate system state at time step $t$. 

We reconstruct the latent phase-space structure by applying time-delay embedding to each variable separately. For the $v$-th variable, the reconstructed state at time $t$ is defined as
\begin{equation}
\begin{split}
z_t^{(v)} =
\bigl[
&\tilde{x}^{(v)}(t-\tau_0),
\tilde{x}^{(v)}(t-\tau_1), \\
&\dots,
\tilde{x}^{(v)}(t-\tau_{m-1})
\bigr]
\in \mathbb{R}^{m},
\end{split}
\end{equation}
where $m$ is the embedding dimension, $\{\tau_i\}_{i=0}^{m-1}$ are learnable delays.

Since the delays are continuous learnable parameters, we compute the delayed value with a differentiable interpolation kernel:
\begin{equation}
	\tilde{x}^{(v)}(t-\tau_i)
	=
	\sum_{j=1}^{T} x_j^{(v)} \, \mathcal{K}(j,\, t-\tau_i),
\end{equation}
where $\mathcal{K}(\cdot,\cdot)$ denotes a differentiable linear interpolation kernel. The reconstructed representations of all variables are then stacked as $Z = \bigl[ z^{(1)}, z^{(2)}, \dots, z^{(V)} \bigr] \in \mathbb{R}^{T \times V \times m}.$ In this way, the model adaptively extracts delay-coordinate patterns that are useful for forecasting, while avoiding fixed offline choices of delay parameters.\\
\textbf{Feature Fusion.}
To obtain a compact phase-space representation for downstream forecasting, we first project the reconstructed attractor-aware states into a latent space:
\begin{equation}
    H_{\mathrm{phase}} = \mathrm{Linear}(Z),
\end{equation}
where $H_{\mathrm{phase}} \in \mathbb{R}^{T \times V \times {d_m}}$ denotes the latent phase-space representation, and ${d_m}$ is the channel-token dimension.

Chaotic time series often exhibit complex nonstationary behavior, in which local statistical patterns vary over time. To retain awareness of the system's global dynamical regime, we further extract a chaos-aware statistics vector from the historical observation:
\begin{equation}
    c = F_{\mathrm{chaos}}(X),
\end{equation}
where $c \in \mathbb{R}^{k}$ contains system-level dynamical descriptors such as complexity, kurtosis, and spectral entropy.

These statistics are then mapped to Feature-wise Linear Modulation (FiLM\cite{perez2018film})
parameters through a Multi-Layer Perceptron (MLP):
\begin{equation}
    [\gamma,\beta] = \mathrm{MLP}(c),
\end{equation}
where $\gamma,\beta \in \mathbb{R}^{d_m}$ denote the channel-wise scale and shift parameters, respectively, and are broadcast along the temporal dimension. The final modulated representation is computed as
\begin{equation}
    H_{\mathrm{fused}} = (1+\gamma)\odot H_{\mathrm{phase}} + \beta,
\end{equation}
where $\odot$ denotes element-wise multiplication. In this way, chaos-aware statistics are used to calibrate the latent phase-space representation, making it more sensitive to the underlying dynamical regime before patch tokenization for LLM-based forecasting.\\
\textbf{Patch Tokenization.}
We divide the fused latent sequence $H_{\mathrm{fused}} \in \mathbb{R}^{T \times V \times d_m}$ into overlapping temporal patches. Following the implementation, we first apply replication padding of length $S$ to the end of the sequence, and then extract $N$ patches of length $P$ with stride $S$, where
\begin{equation}
    N=\left\lfloor \frac{T+S-P}{S} \right\rfloor + 1
    = \left\lfloor \frac{T-P}{S} \right\rfloor + 2.
\end{equation}
Denoting the \(i\)-th patch of the \(v\)-th variable by
\(P_i^{(v)} \in \mathbb{R}^{P \times d_m}\), each patch is flattened and projected into a latent token:
\begin{equation}
    s_i^{(v)} = W_e\,\mathrm{vec}(P_i^{(v)}) + b_e,
    \qquad
    s_i^{(v)} \in \mathbb{R}^{d_m}.
\end{equation}
where \(W_e \in \mathbb{R}^{d_m \times P d_m}\) and \(b_e \in \mathbb{R}^{d_m}\) are learnable projection parameters. This operation maps each flattened patch into the latent feature space, yielding the token sequence \(S \in \mathbb{R}^{V \times N \times d_m}\).
\subsection{Dual-Branch Dynamics Modeling}
\noindent \textbf{LLM-Powered Temporal Modeling.} To harness the generalization capability of pretrained LLMs (e.g., GPT, LLaMA, Qwen series) for numerical time series, we align the reconstructed phase space features with the LLM's semantic space through reprogramming and prompt engineering~\cite{jin2024time}.\par
To provide the frozen LLM with a compact description of the current system, we construct a dynamics-aware prompt from the input series $X$. Specifically, we first extract a compact system descriptor $prompt = F_{\mathrm{stat}}({X}),$ where $F_{\mathrm{stat}}(\cdot)$ summarizes the current system dynamics using statistical and structural descriptors such as trend, energy, and morphology. The prompt design is presented in Figure~\ref{fig:framework}. The descriptor is then converted into a textual prompt and embedded into the LLM space, yielding prompt embeddings $E_{\mathrm{prompt}} \in \mathbb{R}^{(BV) \times L_p \times d_{\mathrm{llm}}},$ where $B$ and $L_p$ denote the batch size and prompt length, $d_{\mathrm{llm}}$ is the hidden dimension of the frozen LLM.

Since the patch-wise time-series tokens and the textual token space of the LLM are heterogeneous, we further employ a lightweight reprogramming module to align the token sequence with the LLM input space. Specifically, we use the patch tokens as queries and a compact set of mapped LLM vocabulary embeddings as keys and values. Let
\(S^{\flat} \in \mathbb{R}^{(BV) \times N \times d_m}
\)
denote the variable-wise patch token sequence, and let
\(
E_{\mathrm{vocab}} \in \mathbb{R}^{M \times d_{\mathrm{llm}}}
\)
denote the mapped source embeddings derived from the frozen LLM vocabulary, where \(M\) is the number of source tokens. The reprogramming module first computes \(Q = S^{\flat} W_Q,
K = E_{\mathrm{vocab}} W_K,
V = E_{\mathrm{vocab}} W_V.\)
Then, cross-attention is applied to obtain LLM-aligned time-series tokens:
\(R =
\operatorname{Proj}_{\mathrm{out}}(
\operatorname{Softmax}
(
\frac{QK^\top}{\sqrt{d_k}}
)V
),
R \in \mathbb{R}^{(BV) \times N \times d_{\mathrm{llm}}}.\)
The prompt embeddings and the reprogrammed tokens are then concatenated along the token dimension to form the LLM input
$I = [\,E_{\mathrm{prompt}} ; R\,] \in \mathbb{R}^{(BV) \times(L_p+N)\times d_{\mathrm{llm}}},
$ where $[\,\cdot\,;\,\cdot\,]$ denotes concatenation along the token axis. This input $I$ is then fed into the frozen LLM backbone to obtain the hidden states $H_{\mathrm{llm}}.$ We retain the hidden states corresponding to the time-series token positions, and project them into the latent forecasting feature space, yielding the following
\begin{equation}
F_{\text{main}} = \text{Flatten} \left( \text{Proj}_{d_f} \left( H_{\text{llm}}[:, L_p + 1 : L_p + N] \right) \right)
\end{equation}
where $F_{\mathrm{main}}$ represents the principal evolution patterns learned by the LLM-powered branch, and $d_f$ denotes the dimension of the flattened forecasting feature.\\
\textbf{Auxiliary Channel Interaction Modeling.}
In multivariate chaotic systems, observed variables are often strongly coupled. While the LLM-powered branch effectively models long-range time dependencies, it does not account for interactions between variables. Thus, we introduce an auxiliary channel interaction branch to capture the complementary dependency structure between variables.\par 
Given the patch-wise token sequence $S$, we first reshape it into a channel-aware representation $U = \mathrm{Reshape}(S) \in \mathbb{R}^{N \times V \times d_m}$, where $V$ is the number of variables and $d_m$ is the channel embedding dimension. This operation explicitly preserves the variable dimension for inter-variable interaction modeling. Based on $U$, we perform channel-wise self-attention at each temporal token position to model inter-variable dependency. For notational simplicity, the following $B$ is omitted. Specifically, for each patch index $n$, the variable-wise token set $U_n \in \mathbb{R}^{V \times d_m}$ is treated as a short sequence over channels, and multi-head self-attention is applied along the channel dimension:
\begin{equation}
    \widetilde{U}_n = \text{Softmax}\left( \frac{Q_n K_n^\top}{\sqrt{d_k}} \right) V_n,
\end{equation}
where $Q_n = U_n W_Q$, $K_n = U_n W_K$, and $V_n = U_n W_V$. The mixed representation is further refined through residual normalization and a lightweight gating function, yielding $U_n^{\text{mix}}$. The resulting interaction-aware tokens are then processed by an auxiliary encoder and projected into the forecasting feature space:
\begin{equation}
    F_{\mathrm{aux}}
    =
    \operatorname{Flatten}\!\left(
    \operatorname{Proj}_{\mathrm{aux}}\!\left(
    \operatorname{Enc}_{\mathrm{aux}}^{\mathrm{MLP}}
    \left(U^{\mathrm{mix}}\right)
    \right)
    \right).
\end{equation}
In this way, the auxiliary interaction branch explicitly models evolving variable dependencies and provides interaction features for the subsequent weighted fusion module.
\subsection{Weighted Fusion Forecasting with Joint Loss Function}
\noindent \textbf{Similarity-Aware Auxiliary Correction.} To measure the consistency between the main temporal representation and the auxiliary interaction features, we directly compute their similarity in the latent forecasting feature space. Let $F_{\mathrm{main}}, F_{\mathrm{aux}} \in \mathbb{R}^{V \times d_f}$ denote the main and auxiliary features. We first evaluate their cosine similarity:
$c_{\mathrm{sim}}^{(v)}
    =
    \frac{
        \left\langle F_{\mathrm{main}}^{(v)},
        F_{\mathrm{aux}}^{(v)}\right\rangle
    }{
        \left\|F_{\mathrm{main}}^{(v)}\right\|_2
        \left\|F_{\mathrm{aux}}^{(v)}\right\|_2
    },
    \quad v=1,\ldots,V.$
This variable-wise similarity score reflects the overall alignment between the dominant temporal evolution captured by the main branch and the interaction-aware auxiliary features.

Based on this similarity cue, we construct a similarity-aware gating input by concatenating the main feature, the auxiliary feature, their absolute difference, and the cosine similarity score:
\begin{equation}
	z_{\mathrm{sim}} =
	\left[
	F_{\mathrm{main}};\,
	F_{\mathrm{aux}};\,
	|F_{\mathrm{main}} - F_{\mathrm{aux}}|;\,
	c_{\mathrm{sim}}
	\right].
\end{equation}
The similarity gate is then obtained through a gating network:
\begin{equation}
	g_{\mathrm{sim}} =
	\sigma\!\left(
	\mathrm{MLP}_{\mathrm{sim}}(z_{\mathrm{sim}})
	\right),
\end{equation}
where $\sigma(\cdot)$ denotes the sigmoid function. Since $g_{\mathrm{sim}} \in \mathbb{R}^{V \times d_f}$, it provides a feature-wise reliability modulation over the auxiliary branch. The auxiliary correction is therefore written as
\begin{equation}
	F_{\mathrm{aux}}^{'} = g_{\mathrm{sim}} \odot F_{\mathrm{aux}}.
\end{equation}
where $\odot$ denotes element-wise multiplication.\\
\textbf{Adaptive Fusion Gate.} To further suppress unreliable auxiliary corrections, we introduce a reliability-aware gate to adaptively control the contribution of the refined auxiliary branch. Let $F_{\mathrm{aux}}^{'}$ denote the similarity-refined feature. The gating input is constructed by feature-wise concatenation:
\begin{equation}
    z_{\mathrm{adp}}
    =
    \operatorname{Concat}\!\left(
    F_{\mathrm{main}},
    F'_{\mathrm{aux}},
    \left|F_{\mathrm{main}}-F'_{\mathrm{aux}}\right|
    \right)
    \in\mathbb{R}^{V\times 3d_f}.
\end{equation}
The reliability-aware gate is then computed by $\boldsymbol{\alpha}=\sigma\!\left(\mathrm{MLP}_{\mathrm{adp}}(\mathbf{z}_{\mathrm{adp}})\right)$, where $\sigma(\cdot)$ denotes the sigmoid function. Since $\alpha \in \mathbb{R}^{V \times d_f}$, it acts as a reliability estimator, adaptively determining how much auxiliary information should be injected into the main forecasting branch.

To further transform the refined auxiliary representation into a correction term that is compatible with the main feature space, we apply the auxiliary delta head:
\begin{equation}
    \Delta F
    =
    \mathrm{Head}_{\Delta}\!\left(F'_{\mathrm{aux}}\right),
\end{equation}
The fused representation is then formulated as
\begin{equation}
	F_{\mathrm{fusion}} =
	F_{\mathrm{main}} + \alpha \odot \Delta F.
\end{equation}
In this way, PAC-LLM uses the reliability-aware gate to suppress unreliable auxiliary corrections while preserving beneficial interaction-aware information for final forecasting.\\
\textbf{Forecasting.} $F_{\mathrm{fusion}}$ denotes the fused representation generated by the adaptive fusion module. Based on this fused feature stream, we employ a residual forecasting head to generate the future trajectory:
\begin{equation}
	\hat{\mathbf{Y}} =
	\mathrm{Proj}_{\mathrm{head}}\!\left(
	\mathrm{MLP}(F_{\mathrm{fusion}})
	\right)
	+
	\mathrm{Proj}_{\mathrm{res}}(F_{\mathrm{fusion}}).
\end{equation}
Here, $\mathrm{Proj}_{\mathrm{head}}(\cdot)$ and $\mathrm{Proj}_{\mathrm{res}}(\cdot)$ denote learnable projection layers for the residual forecasting head. The output $\hat{\mathbf{Y}}$ is finally reshaped into $\mathbb{R}^{B \times H \times V}$, where $H$ denotes the prediction horizon. In this way, the forecasting head acts as the final projection module that translates the fused latent representation into the predicted future trajectory of the chaotic system.\\
\textbf{Training Objective.}
The fused forecast $\hat{\mathbf{Y}}$ is directly supervised by the ground-truth future trajectory $\mathbf{Y}$. To constrain the auxiliary branch, an independent auxiliary forecasting head maps the pre-fusion representation $F'_{\mathrm{aux}}$ to
$\hat{\mathbf{Y}}_{\mathrm{aux}}
=\mathrm{Head}_{\mathrm{aux}}(F'_{\mathrm{aux}})$.
The two forecasting losses are defined as
\begin{equation}
    \mathcal{L}_{\mathrm{fusion}}
    =\frac{1}{BHV}
    \left\|\hat{\mathbf{Y}}-\mathbf{Y}\right\|_F^2,
    \qquad
    \mathcal{L}_{\mathrm{aux}}
    =\frac{1}{BHV}
    \left\|\hat{\mathbf{Y}}_{\mathrm{aux}}-\mathbf{Y}\right\|_F^2,
\end{equation}
The overall training objective is a weighted combination of
the two terms: $\mathcal{L}
    =\lambda_f\mathcal{L}_{\mathrm{fusion}}
    +(1-\lambda_f)\mathcal{L}_{\mathrm{aux}}$, where $\lambda_f$ is a balancing coefficient.

\section{Experiment}
\begin{table*}[t]
\centering
\caption{Performance comparison of different models across various chaotic systems and evaluation metrics. sMAPE@1, sMAPE@4, and sMAPE@10 correspond to predictions at 30, 120, and 300 steps (approximately 1, 4, and 10 Lyapunov times). Metrics marked with ($\uparrow$) indicate higher is better, while ($\downarrow$) indicates lower is better. The best performance of each metric is marked in \textbf{bold}, and the second-best performance is \underline{underlined}.}
\label{tab:model_comparison}
\resizebox{\linewidth}{!}{%
\begin{tabular}{l c c c c c c c c c c c}
\toprule
System & \diagbox{Metric}{Model} & Ours & TimeLLM & LLMMixer & CrossFormer & NBEATS & TimesNet & nVAR & ESN & LSTNet & Panda \\
\midrule

\multirow{6}{*}{Lorenz63 (3D)}
& VPT ($\uparrow$)
& \textbf{4.57} & 0.67 & 0.23 & 1.83 & 1.20 & \underline{3.40} & 1.17 & 0.13 & 1.97 & 0.02 \\
& sMAPE@1 ($\downarrow$)
& \textbf{8.14} & 36.40 & 53.00 & 19.02 & 23.11 & \underline{12.55} & 33.39 & 81.28 & 18.47 & 137.21 \\
& sMAPE@4 ($\downarrow$)
& \textbf{23.93} & 95.76 & 111.97 & 64.33 & 72.29 & \underline{38.69} & 115.27 & 119.78 & 76.65 & 142.20 \\
& sMAPE@10 ($\downarrow$)
& \textbf{81.51} & 125.79 & 132.55 & 107.27 & 112.92 & \underline{93.31} & 144.49 & 132.94 & 110.72 & 142.65 \\
& $D_{\text{frac}}$ ($\downarrow$)
& \textbf{0.060} & 0.240 & 0.189 & 0.077 & 0.292 & 0.077 & 0.141 & 0.341 & \underline{0.068} & 0.108 \\
& $D_{\text{stsp}}$ ($\downarrow$)
& \textbf{0.043} & 0.520 & 0.683 & 0.066 & 0.149 & 0.070 & 1.284 & 0.503 & \underline{0.052} & 0.653 \\

\midrule

\multirow{6}{*}{Rossler (3D)}
& VPT ($\uparrow$)
& \textbf{9.53} & 0.83 & 0.33 & \underline{7.80} & 7.20 & 7.00 & 0.13 & 0.02 & 4.51 & 0.00 \\
& sMAPE@1 ($\downarrow$)
& \textbf{2.24} & 25.54 & 36.00 & 7.81 & 5.34 & \underline{4.06} & 145.57 & 129.49 & 9.59 & 141.36 \\
& sMAPE@4 ($\downarrow$)
& \textbf{5.58} & 66.22 & 73.39 & 15.46 & \underline{15.20} & 17.37 & 167.35 & 143.37 & 28.29 & 130.85 \\
& sMAPE@10 ($\downarrow$)
& \textbf{19.68} & 105.59 & 106.28 & \underline{31.88} & 41.93 & 39.43 & 171.89 & 146.13 & 67.02 & 133.38 \\
& $D_{\text{frac}}$ ($\downarrow$)
& \textbf{0.064} & 0.220 & 0.327 & 0.083 & 0.122 & \underline{0.079} & 1223.331 & 0.352 & 0.127 & 0.175 \\
& $D_{\text{stsp}}$ ($\downarrow$)
& \textbf{0.004} & 0.459 & 0.654 & \underline{0.007} & 0.027 & 0.030 & 5.478 & 3.226 & 0.054 & 0.498 \\
\midrule

\multirow{6}{*}{Chua (3D)}
& VPT ($\uparrow$)
& \textbf{8.50} & 0.63 & 0.73 & 6.97 & 3.73 & 2.43 & 0.06 & 0.77 & \underline{8.29} & 0.00 \\
& sMAPE@1 ($\downarrow$)
& \textbf{4.63} & 47.83 & 41.62 & 5.60 & 6.34 & 16.56 & 37.04 & 51.81 & \underline{4.86} & 123.56 \\
& sMAPE@4 ($\downarrow$)
& \textbf{9.83} & 92.50 & 92.22 & 17.84 & 36.48 & 59.90 & 100.99 & 92.98 & \underline{13.42} & 125.47 \\
& sMAPE@10 ($\downarrow$)
& \textbf{31.73} & 121.67 & 125.86 & 52.04 & 80.72 & 103.60 & 120.45 & 117.06 & \underline{38.85} & 138.86 \\
& $D_{\text{frac}}$ ($\downarrow$)
& \textbf{0.051} & 0.104 & 0.150 & 0.078 & 0.130 & 0.110 & 0.138 & 0.149 & \underline{0.061} & 0.156 \\
& $D_{\text{stsp}}$ ($\downarrow$)
& \textbf{0.037} & 1.081 & 1.153 & 0.070 & 0.241 & 0.603 & 0.604 & 0.189 & \underline{0.045} & 1.621 \\
\midrule

\multirow{6}{*}{Lorenz96 (8D)}
& VPT ($\uparrow$)
& \textbf{1.77} & 0.12 & 0.19 & 0.45 & 0.34 & 0.17 & \underline{1.53} & 0.11 & 0.98 & 0.00 \\
& sMAPE@1 ($\downarrow$)
& \underline{23.06} & 103.64 & 96.18 & 48.15 & 70.80 & 51.66 & \textbf{10.50} & 92.60 & 32.28 & 137.74 \\
& sMAPE@4 ($\downarrow$)
& \textbf{46.99} & 138.52 & 135.19 & 105.40 & 124.18 & 109.77 & 117.85 & 129.74 & \underline{89.69} & 145.39 \\
& sMAPE@10 ($\downarrow$)
& \textbf{92.25} & 142.59 & 143.87 & 126.20 & 138.49 & 132.03 & 151.45 & 138.07 & \underline{121.59} & 144.21 \\
& $D_{\text{frac}}$ ($\downarrow$)
& \textbf{0.243} & 1.316 & 0.694 & 0.503 & 0.398 & 0.608 & 0.988 & 0.390 & \underline{0.312} & 0.834 \\
& $D_{\text{stsp}}$ ($\downarrow$)
& \textbf{0.434} & 3.182 & 3.600 & 0.802 & 1.159 & 2.317 & 4.602 & 1.373 & \underline{0.615} & 3.787 \\

\bottomrule
\end{tabular}%
}
\end{table*}
\subsection{Experimental Setup}
\noindent\textbf{Datasets.} We evaluate our model on multiple chaotic datasets generated from the dysts library~\cite{gilpin2021chaos}, including four representative systems: Lorenz63, Rossler, Chua, and Lorenz96. We use a fourth-order Runge-Kutta numerical integrator to generate trajectories for different datasets with an integration step of $\Delta t = 0.001$. For the Lorenz96 system, an 8-dimensional setup is adopted in our experiments. To make the temporal resolution comparable across systems with different intrinsic time scales, we compute the Lyapunov time of each system and downsample the raw trajectories to approximately 30 points per Lyapunov time. Before data collection, the initial transient corresponding to 5 Lyapunov times is discarded. For each system, we generate one long training trajectory with 30,000 points and multiple test trajectories from unseen initial conditions sampled on the attractor. To evaluate the robustness of PAC-LLM on real-world noisy datasets, we further conduct experiments on the IBM Double Pendulum Chaotic Dataset~\cite{asseman2018learning}. The dataset contains 21 independent physical double-pendulum experiments recorded at approximately 400 fps. From the tracked marker trajectories, we construct a four-dimensional state consisting of two angular positions and their corresponding angular velocities. The sequences are downsampled by a factor of three to approximately 133.33 Hz.

\noindent\textbf{Baselines.} We compare our method with a diverse set of competitive time series forecasting baselines from six categories: (1) dynamical models: NVAR~\cite{gauthier2021next}, ESN~\cite{jaeger2004harnessing}. (2) LLM-based methods: Time-LLM~\cite{jin2024time}, LLM-Mixer~\cite{kowsher2025llm}. (3) Transformer-based models: Crossformer~\cite{zhang2023crossformer}. (4) CNN/RNN-based methods: TimesNet~\cite{wu2022timesnet} and LSTNet~\cite{lai2018modeling}. (5) MLP-based method: NBEATS~\cite{oreshkin2019n}. (6) foundation model: Panda~\cite{lai2025panda}. In particular, LSTNet is included as a representative deep forecasting baseline that has been commonly adopted in chaotic time-series prediction studies~\cite{valle2025forecasting}.\par
\noindent\textbf{Experimental Protocol.} We employ a uniform autoregressive rolling prediction protocol for all forecasting experiments. The input window length is fixed at 30, and during training, predictions are made for the next 30 steps. During testing, autoregressive forecasts are evaluated at $H\in\{30,120,300\}$ up to 300 steps, where $H=30$ measures short-term accuracy and $H=120,300$ assess long-horizon rollout performance. Input-length sensitivity is additionally evaluated by varying only the observation window while aligning all forecast origins using an anchor length of 60. For the synthetic chaotic systems, we use a single-train-multi-test partitioning approach. Each dataset contains one long training trajectory and several test trajectories with different initial values. The training trajectory is further divided into training and validation sets in chronological order, with a ratio of 0.85:0.15. For the noisy real-world double-pendulum dataset, the 21 experimental trajectories are split into 15/3/3 trajectories for training, validation, and testing.

We use the first two layers of GPT-2~\cite{radford2019language} with a hidden size of 768 as the default language backbone. To assess backbone dependence, we additionally replace GPT-2 with Qwen2.5-3B~\cite{hui2024qwen2} and Llama-3.2-3B~\cite{grattafiori2024llama}, while keeping the remaining PAC-LLM architecture and training protocol unchanged. We train the model using Adam with learning rate $5 \times {10^{ - 5}}$, weight decay ${10^{ - 4}}$, batch size 32, dropout 0.1 and at most 150 epochs. All experiments are conducted on an NVIDIA RTX 4080 SUPER-based virtual GPU with a 32 \,GB memory profile. For the delay-embedding dimension $m$, we use $m=3$ for Lorenz63, Rossler, and Chua, $m=8$ for the higher-dimensional Lorenz96 system, and $m=5$ for the double-pendulum system. Representative hyperparameter settings include a patch size of $P=8$ and a stride of $S=4$, loss-balancing coefficient $\lambda_f=0.9$, and hidden dimension $d_m=32$. The forecasting feature dimension $d_f$ is set to 128 for most datasets, providing a lightweight yet expressive latent representation.

\noindent\textbf{Evaluation Metrics.} We evaluate long-term chaotic forecasting from multiple complementary aspects, including point-wise accuracy, valid prediction time (VPT), and dynamical consistency of autoregressive rollouts. We use mean squared error (MSE), mean absolute error (MAE), and symmetric mean absolute percentage error (sMAPE) to measure point-wise forecast deviations between predicted and true states over different forecasting horizons. VPT quantifies how long the predicted trajectory remains reliable before the error exceeds a predefined threshold. VPT uses the same cumulative-sMAPE threshold of 30\% for all systems. It is reported in Lyapunov-time units for the synthetic systems and in seconds for the experimental double-pendulum system. To further assess whether long rollouts preserve intrinsic dynamics of the system, we report the correlation dimension discrepancy \(D_{\mathrm{frac}}\) and Kullback–Leibler Divergence between attractors \(D_{\mathrm{stsp}}\), which characterize discrepancies in fractal geometry and state-space distribution.

\subsection{Main Results}
We compare our model with several representative baseline models to evaluate its effectiveness in predicting chaotic systems. Table~\ref{tab:model_comparison} summarizes the experimental results on four datasets, where sMAPE@1 reflects short-term predictive ability, sMAPE@4 and sMAPE@10 reflect prediction robustness under long-term error accumulation. Long-term dynamics preservation ability is measured by \(D_{\mathrm{frac}}\) and \(D_{\mathrm{stsp}}\). Based on the experimental results, the following observation can be drawn:\\
\noindent\textbf{\ding{172} Strong overall performance across systems and metrics.} Among all comparative experiments, our method achieves the best performance in 23 cases and ranks second in the remaining one case. The improvements are not limited to a single type of metric: sMAPE at different horizons reflects trajectory-tracking accuracy from short-term to long-term prediction, while \(D_{\mathrm{frac}}\) and \(D_{\mathrm{stsp}}\) evaluate whether long rollouts preserve the geometric and distributional properties of chaotic attractors.

\noindent\textbf{\ding{173} Greater gains at long horizons.}
Although the performance gap is already visible at (H=30), it becomes more pronounced at (H=120) and (H=300), where autoregressive error accumulation increasingly challenges the model. The consistent improvements in horizon-dependent metrics, such as sMAPE and VPT, indicate that our method can follow the true trajectory for a longer predictable interval before divergence. In other words, the proposed framework is not only effective for short-term fitting, but also better sustains trajectory-tracking quality under long-horizon rollout.

\noindent\textbf{\ding{174} Better dynamical reconstruction beyond point-wise accuracy.}
The advantage of our method is also reflected in attractor-related metrics, including \(D_{\mathrm{frac}}\) and \(D_{\mathrm{stsp}}\), which are especially important for long-term chaotic forecasting. A lower \(D_{\mathrm{frac}}\) indicates that the reconstructed trajectory better preserves the fractal geometric complexity of the underlying attractor, while a lower \(D_{\mathrm{stsp}}\) suggests a closer state-space distribution between predicted and true trajectories. These results show that our method does not merely reduce point-wise forecast errors, but also improves the preservation of global dynamical structures during long autoregressive prediction.

\begin{figure*}[t]
    \centering
    \includegraphics[width=\textwidth]{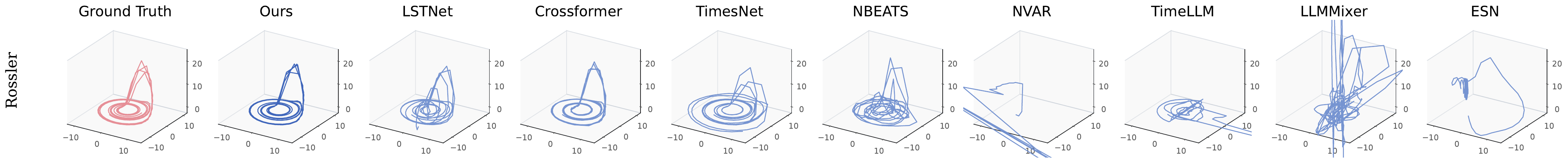}
    
    
    \includegraphics[width=\textwidth]{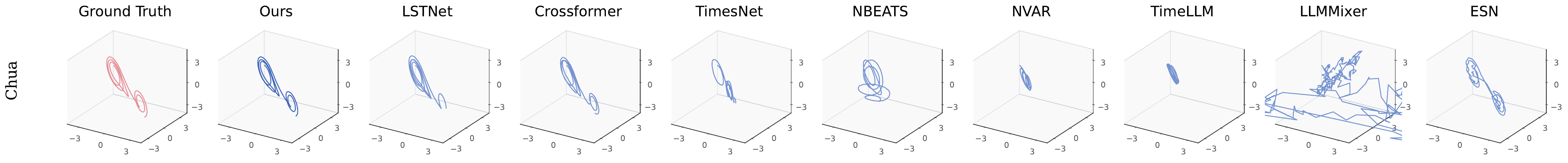}
    \caption{Phase-space comparison of long-term forecasting results on the Rossler and Chua systems..}
    \label{fig:phase_compare}
\end{figure*}

\begin{figure*}[t]
    \centering
    \includegraphics[width=0.85\textwidth]{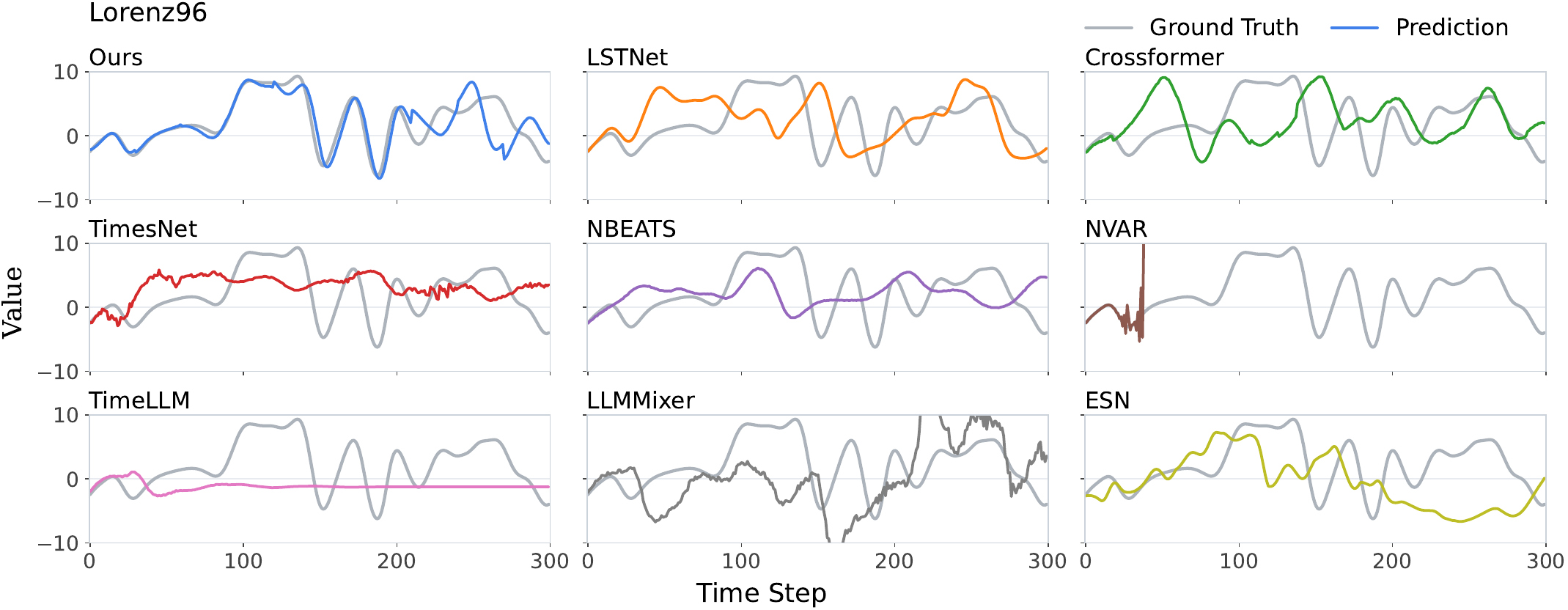}
    \caption{Long-term rollout forecasting results on the Lorenz96 system. The colored solid lines represent the predicted trajectories of different models, while the gray solid line denotes the ground truth.}
    \label{timeseries}
\end{figure*}

Furthermore, Figure~\ref{fig:phase_compare} shows the phase space structure of the Rossler and Chua systems, demonstrating the superior ability of our method to reconstruct the complex geometry of attractors. In Figure~\ref{timeseries}, we visualize predicted trajectories over
the next 300 steps (i.e., 10 Lyapunov times) of Lorenz96 (8D), which is the most complex chaotic system in our experiment. We can observe that the predicted trajectory from PAC-LLM is much closer to the ground truth compared to other baselines. 

\subsection{Ablation Study} 
\begin{table}[t]
\centering
\caption{Ablation study on Lorenz63 dataset. The best performance of each metric is marked in \textbf{bold}.}
\label{tab:ablation}
\renewcommand{\arraystretch}{1.15}
\setlength{\tabcolsep}{4pt}

\begin{tabular}{l|ccc}
\toprule
Type & \multicolumn{3}{c}{MSE ($\downarrow$) / MAE ($\downarrow$)} \\
\cmidrule(lr){2-4}
& 30 steps & 120 steps & 300 steps \\
\midrule

\multicolumn{4}{l}{\textbf{Temporal feature module}} \\[-2pt]

w/o TF
& 0.032 / 0.065
& 0.461 / 0.326
& 1.202 / 0.723 \\

\quad w/o Takens
& 0.017 / 0.038
& 0.259 / 0.208
& 1.100 / 0.655 \\

\quad w/o Chaos
& 0.014 / 0.032
& 0.227 / 0.180
& 1.101 / 0.638 \\

\addlinespace[2pt]
\midrule

\multicolumn{4}{l}{\textbf{Auxiliary and fusion components}} \\[-2pt]

w/o Aux
& 0.022 / 0.072
& 0.534 / 0.373
& 1.295 / 0.767 \\

w/o WF
& \textbf{0.009} / 0.035
& 0.231 / 0.195
& 1.140 / 0.667 \\

w/o Aux Loss
& 0.011 / 0.028
& 0.148 / 0.133
& 1.039 / 0.604 \\

\addlinespace[2pt]
\midrule

\multicolumn{4}{l}{\textbf{LLM input}} \\[-2pt]

w/o Prompt
& 0.012 / 0.029
& 0.190 / 0.160
& 1.090 / 0.632 \\

\midrule

Full model
& 0.012 / \textbf{0.027}
& \textbf{0.139} / \textbf{0.127}
& \textbf{1.021} / \textbf{0.593} \\

\bottomrule
\end{tabular}
\end{table}
To verify the effectiveness of each key module of the model, we further designed ablation experiments to examine the roles of the 1). Phase-space-aware representation module, 2). auxiliary branches, 3). weighted fusion mechanisms, and 4). auxiliary supervision objectives. Specifically, the following variations are included:
\begin{itemize}
    \item \textbf{w/o TF}: Both the Learnable Delay Embedding component and the FiLM fusion component are removed. We additionally consider \textbf{w/o Takens}, which removes only the Learnable Delay Embedding, and \textbf{w/o Chaos}, which removes only the chaos-aware statistical features and the corresponding FiLM modulation.
    \item \textbf{w/o Aux}: The auxiliary branch is removed, and the LLM-based main branch is retained to generate
the prediction independently.
    \item \textbf{w/o WF}: The adaptive weighted fusion module is removed, the latent
representations are instead concatenated and projected directly to the
main prediction head, eliminating the adaptive weighting of the
auxiliary correction.
    \item \textbf{w/o Aux Loss}: The auxiliary supervision is removed, while the auxiliary branch and the fusion mechanism are retained. The training objective contains only the main forecasting loss.
    \item \textbf{w/o Prompt}: The textual prompt is removed, while the
numerical patches are still reprogrammed and processed by the pretrained
LLM backbone. 
\end{itemize}
The results on Lorenz63 are shown in Table~\ref{tab:ablation}, from which several observations can be drawn:\\
\noindent\textbf{\ding{172} The auxiliary branch provides essential complementary dynamics.}
Removing the auxiliary branch (\textit{w/o Aux}) causes the largest degradation, increasing the average MSE and MAE by 57.9\% and 62.2\%, respectively. This confirms that the LLM-based branch alone is insufficient for robust long-horizon chaotic forecasting.\\
\noindent\textbf{\ding{173} Phase-space-aware temporal features are important.}
Removing the complete temporal feature module (\textit{w/o TF}) increases the average MSE and MAE by 44.6\% and 49.1\%. Further ablations show that removing Takens embedding or chaos-aware modulation individually also degrades performance, confirming that both components contribute to the dynamics-aware representation.\\
\noindent\textbf{\ding{174} Fusion quality matters.}
Removing weighted fusion (\textit{w/o WF}) increases the average MSE and MAE by 17.8\% and 20.1\%, with a more evident degradation at longer horizons, demonstrating the importance of adaptively integrating auxiliary corrections. Auxiliary supervision provides a modest but consistent overall benefit.\\
\noindent\textbf{\ding{175} Prompt information provides additional gains.}
Removing the prompt (\textit{w/o Prompt}) increases the average MSE and MAE, indicating that semantic prompt information complements the numerical representations. However, results in Table~\ref{tab:ablation} indicate that it is not the primary source of performance gains.

\subsection{Analysis of the LLM Component}
\noindent\textbf{LLM Contribution Analysis.} 
To isolate the contribution of the pretrained LLM backbone, PAC-Transformer replaces the pretrained GPT-2 blocks with a randomly initialized Transformer while retaining the LLM-style input pathway. PAC-Numeric-Transformer further removes prompt construction and reprogramming and directly operates on numerical representations. PAC-Transformer uses a two-layer Transformer with a hidden size of 768, matching the depth and representation dimension of the GPT-2 backbone. PAC-Numeric-Transformer processes numerical patch representations using a two-layer Transformer with a hidden size of 768. As shown in Table~\ref{tab:llm_backbone_control}, PAC-LLM generally outperforms the two non-LLM variants across the reported metrics. This suggests that pretrained LLM representations provide complementary benefits beyond the dynamics-aware numerical components. The gains therefore cannot be attributed solely to the Transformer architecture or the reprogramming interface.\par
\begin{table}[t]
\centering
\caption{
Controlled backbone replacement study on Lorenz63 and Rossler. The best performance of each metric is marked in \textbf{bold}.
}
\label{tab:llm_backbone_control}
\resizebox{\columnwidth}{!}{
\begin{tabular}{llccc}
\toprule
System & Metric
& PAC-LLM
& PAC-Transformer
& PAC-Numeric-Transformer \\
\midrule

\multirow{4}{*}{Lorenz63}
& VPT ($\uparrow$)
& \textbf{4.57}
& 2.30
& \underline{4.12} \\

& sMAPE@1 ($\downarrow$)
& \textbf{8.14}
& 15.75
& \underline{9.65} \\

& $D_{\mathrm{frac}}$ ($\downarrow$)
& \textbf{0.060}
& 0.123
& \underline{0.068} \\

& $D_{\mathrm{stsp}}$ ($\downarrow$)
& \textbf{0.043}
& 0.088
& \underline{0.051} \\

\midrule

\multirow{4}{*}{Rossler}
& VPT ($\uparrow$)
& \textbf{9.53}
& 8.83
& \underline{9.52} \\

& sMAPE@1 ($\downarrow$)
& \textbf{2.24}
& 3.83
& \underline{2.34} \\

& $D_{\mathrm{frac}}$ ($\downarrow$)
& \textbf{0.064}
& 0.076
& \underline{0.065} \\

& $D_{\mathrm{stsp}}$ ($\downarrow$)
& \underline{0.004}
& 0.010
& \textbf{0.0036} \\

\bottomrule
\end{tabular}
}
\end{table}
\noindent\textbf{LLM Backbone Analysis.} 
To further examine the role of the LLM component, we evaluate PAC-LLM with GPT-2, Qwen2.5-3B, and Llama3.2-3B on the Rossler system. All three backbones can be integrated into PAC-LLM without modifying the overall forecasting architecture. Notably, simply increasing the backbone scale does not consistently improve forecasting performance, while fine-tuning GPT-2 achieves lower long-term errors. These results shown in Table~\ref{tab:llm_backbone_adaptation} suggest that task-specific adaptation can be more beneficial than simply using a larger pretrained backbone.
\begin{table}[t]
\centering
\caption{
LLM backbone and adaptation results on Rossler (MSE/MAE; lower is better).
}
\label{tab:llm_backbone_adaptation}
\resizebox{\columnwidth}{!}{
\begin{tabular}{lccc}
\toprule
Variant & 30-step & 120-step & 300-step \\
\midrule
GPT-2 (frozen)
& 1.27e-4/\textbf{5.71e-3}
& \textbf{1.63e-3}/\textbf{1.69e-2}
& 0.115/0.108 \\

GPT-2 (fine-tuned)
& 1.39e-4/5.94e-3
& 1.72e-3/1.72e-2
& \textbf{0.104}/\textbf{0.103} \\

Qwen2.5-3B
& \textbf{1.26e-4}/6.09e-3
& 2.21e-3/2.01e-2
& 0.133/0.122 \\

Llama-3.2-3B
& 1.90e-4/8.07e-3
& 4.15e-3/2.36e-2
& 0.135/0.127 \\
\bottomrule
\end{tabular}
}
\end{table}
\subsection{Robustness and Real-World Evaluation}
\noindent\textbf{Robustness to Observation Noise.} 
We evaluate PAC-LLM by adding Gaussian noise with different intensities to the observed input sequences while keeping the trained model unchanged. PAC-LLM remains relatively stable under mild perturbations up to $\sigma=10^{-2}$, whereas larger noise levels lead to a clear degradation in both forecasting accuracy and attractor reconstruction. This indicates that PAC-LLM is robust to moderate observation noise but remains sensitive to severe corruption.\par
\begin{figure}[t]
    \centering
    \includegraphics[width=0.46\textwidth]{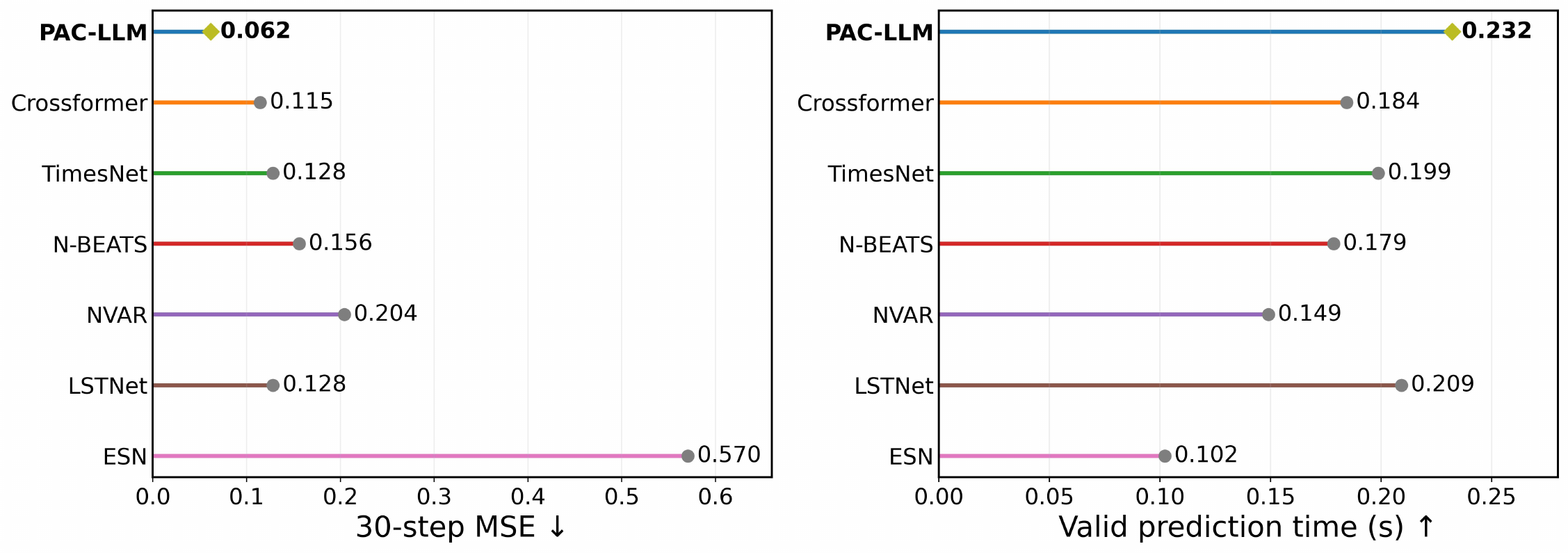}
    \caption{Real-world chaotic forecasting on the experimental double-pendulum dataset. 
             Left: Short-term accuracy. 
             Right: VPT in seconds.}
    \label{fig:double_pendulum}
\end{figure}

\noindent\textbf{Real-World Double-Pendulum Forecasting.}
We further evaluate PAC-LLM on experimentally recorded double-pendulum trajectories. From the tracked markers, we construct $\mathbf{x}_t=[\theta_1,\theta_2,\dot{\theta}_1,\dot{\theta}_2]$, with unwrapped angles and numerically differentiated angular velocities, preserving measurement noise and experimental variability. The sequences are downsampled by a factor of three to approximately $133.33$ Hz ($\Delta t=0.0075$ s). Following the same setting as the synthetic experiments, we evaluate autoregressive rollouts at $H\in\{30,120,300\}$, corresponding to approximately $0.225$, $0.9$, and $2.25$ s. Unlike the synthetic systems, the experimental double pendulum is affected by friction and gradual energy dissipation. Because a single global Lyapunov time is not assumed for these non-stationary experimental trajectories, We report VPT in physical seconds rather than in Lyapunov time units. It can be seen in Figure~\ref{fig:double_pendulum}, PAC-LLM achieves the lowest short-horizon MSE and the longest VPT. This demonstrates the applicability of PAC-LLM to noisy real-world chaotic dynamics. 
\subsection{Extended Analysis}
\noindent\textbf{Efficiency Analysis.} 
Table \ref{tab:efficiency} compares the efficiency of Time-LLM, LLM-Mixer, and our proposed method on the Lorenz63 system, all of which are also based on LLM. It can be observed that PAC-LLM exhibits a superior overall advantage in terms of memory usage and inference speed, thus achieving a better balance between efficiency and prediction performance.
\begin{table}[t]
	\centering
	\caption{Efficiency comparison on Lorenz63. Param. denotes model parameters, Mem. denotes memory usage, and Speed denotes seconds per iteration.}
	\small
	\setlength{\tabcolsep}{5pt}
    \setlength{\textfloatsep}{4pt}
	\renewcommand{\arraystretch}{1.15}
	\begin{tabular}{lccc}
		\toprule
		\textbf{Model} & \textbf{Param. (M)} & \textbf{Mem. (MB)} & \textbf{Speed (s/iter)} \\
		\midrule
		Time-LLM & 52.68 & 3,163 & 0.1110 \\
		LLM-Mixer & \textbf{5.10} & 3,740 & 0.1121 \\
		Ours & 61.08 & \textbf{2,458} & \textbf{0.0898} \\
		\bottomrule
	\end{tabular}
	\label{tab:efficiency}
\end{table}

\begin{figure}[t]
    \centering
    \includegraphics[width=0.47\textwidth]{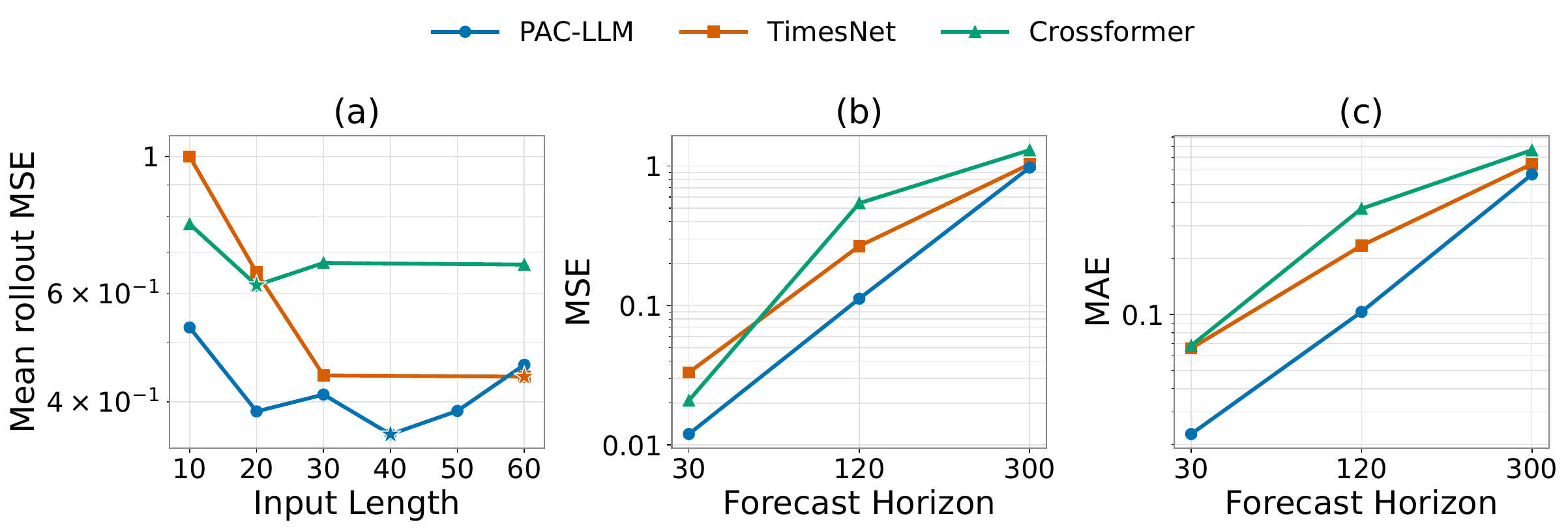}
    \caption{Input-length comparison on Lorenz63. (a) Stars mark the selected best contexts.
    (b)--(c) MSE and MAE at all horizons using each method's selected context.}
    \label{fig:input_length}
\end{figure}

\noindent\textbf{Sensitivity Analysis.}
To ensure that the comparison is not biased toward the short context used by PAC-LLM, we independently tune the input length of each method using only the validation set. All candidate input lengths share the same forecast origins, so that only the amount of observed history changes. As shown in Figure~\ref{fig:input_length}(a), we compare PAC-LLM against two competitive baselines: TimesNet and Crossformer. Their selected context lengths are 60 and 20 respectively, whereas PAC-LLM favors 40. This variation indicates that a longer context is not uniformly beneficial and that the preferred observation length depends on the model architecture. Figure~\ref{fig:input_length}(b)--(c) further reports the test performance obtained with the context selected for each method. PAC-LLM achieves lower MSE and MAE across different horizons, demonstrating that its advantage is retained when every method is allowed to use its own preferred input length.
\par
\begin{figure}[t]
    \centering
    \includegraphics[width=0.47\textwidth]{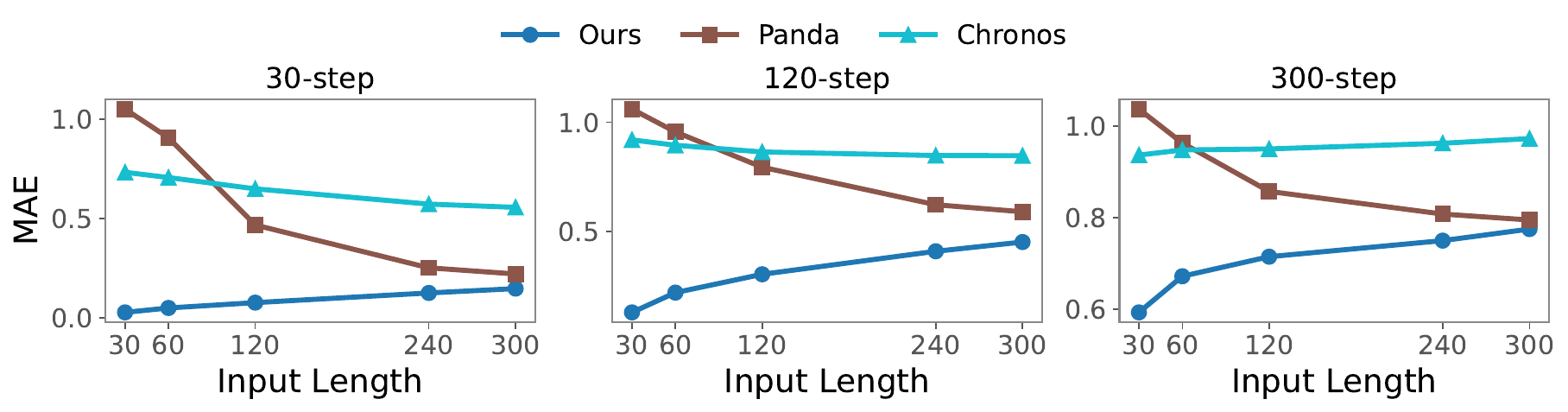}
    
    \includegraphics[width=0.47\textwidth]{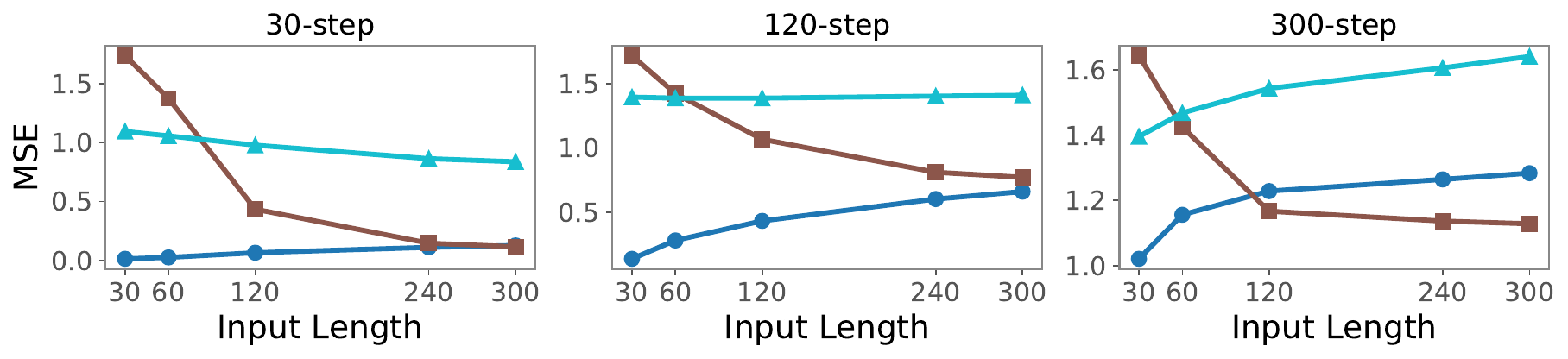}
    \caption{Context-length sensitivity of PAC-LLM and foundation model
baselines on Lorenz63. MAE (top) and MSE (bottom) are reported for the
30-, 120-, and 300-step rollout horizons.}
    \label{fig:input_length_analysis}
\end{figure}

\noindent\textbf{Comparison with Foundation Models.}
Figure~\ref{fig:input_length_analysis} further compares the context-length sensitivity
of PAC-LLM with the foundation-model baselines Panda and Chronos. The curves
show clearly different context-use regimes. PAC-LLM performs best with short
observations, while its error generally increases as increasingly old
measurements are included. In contrast, Panda benefits substantially from
additional context, particularly for long-horizon forecasting. Chronos exhibits a comparatively weak or horizon-dependent response to
longer inputs. This behavior is consistent with the different forecasting
mechanisms: PAC-LLM emphasizes local state reconstruction from recent delayed
observations, while the foundation models can use longer contexts for
sequence-level pattern matching. The intersections show that PAC-LLM has a clear advantage under limited observations, whereas Panda requires substantially longer histories to match or surpass PAC-LLM on only some metrics. This contrast highlights PAC-LLM's effectiveness in the limited observation regime.
\section{Conclusion and Future Work}
This work focuses on long-term chaotic time series forecasting. We first propose a unified setting to facilitate this line of research. Then, we develop PAC-LLM, which incorporates learnable delay embedding and chaos statistics extractor to enhance local dynamical representations, and further integrates an LLM-powered main branch with an auxiliary coupling branch through gated adaptive fusion. Experiments on multiple chaotic systems show that PAC-LLM achieves consistent improvements across point-wise forecasting metrics and attractor-related dynamical metrics, demonstrating its effectiveness in long-term chaotic time series forecasting. Ablation studies further confirm the contributions of phase-space-aware representation learning, auxiliary coupling modeling, adaptive fusion and auxiliary supervision. In future work, we will further explore incorporating physics-informed constraints~\cite{feng2025toward} into PAC-LLM to enhance its generalization capability for forecasting more real-world chaotic systems.


\bibliographystyle{ACM-Reference-Format}
\bibliography{main}

\end{document}